\documentclass[sigconf,authorversion,nonacm]{acmart}
\AtBeginDocument{%
  }

\setcopyright{acmlicensed}
\copyrightyear{2018}
\acmYear{2018}
\acmDOI{XXXXXXX.XXXXXXX}

\acmConference[Conference acronym 'XX]{Make sure to enter the correct
  conference title from your rights confirmation emai}{June 03--05,
  2018}{Woodstock, NY}
\acmISBN{978-1-4503-XXXX-X/18/06}

\begin{document}

\title{LoRE: Logit-Ranked Retriever Ensemble for Enhancing Open-Domain Question Answering}

\author{Saikrishna Sanniboina}
\email{ss235@illinois.edu}
\affiliation{%
  \institution{University of Illinois at Urbana-Champaign}
  \city{Urbana}
  \state{Illinois}
  \country{USA}
}

\author{Shiv Trivedi}
\email{shivvt@illinois.edu}
\affiliation{%
  \institution{University of Illinois at Urbana-Champaign}
  \city{Urbana}
  \state{Illinois}
  \country{USA}
}

\author{Sreenidhi Vijayaraghavan}
\email{sv23@illinois.edu}
\affiliation{%
  \institution{University of Illinois at Urbana-Champaign}
  \city{Urbana}
  \state{Illinois}
  \country{USA}
}


\begin{abstract}
Retrieval-based question answering systems often suffer from positional bias, leading to suboptimal answer generation. We propose LoRE (Logit-Ranked Retriever Ensemble), a novel approach that improves answer accuracy and relevance by mitigating positional bias. LoRE employs an ensemble of diverse retrievers, such as BM25 and sentence transformers with FAISS indexing. A key innovation is a logit-based answer ranking algorithm that combines the logit scores from a large language model (LLM), with the retrieval ranks of the passages. Experimental results on NarrativeQA, SQuAD demonstrate that LoRE significantly outperforms existing retrieval-based methods in terms of exact match and F1 scores. On SQuAD, LoRE achieves 14.5\%, 22.83\%, and 14.95\% improvements over the baselines for ROUGE-L, EM, and F1, respectively. Qualitatively, LoRE generates more relevant and accurate answers, especially for complex queries.
\end{abstract}




\maketitle

\section{Introduction}
Question answering (QA) systems are pivotal in the landscape of natural language processing, designed to decipher user queries and retrieve relevant information from extensive sources such as databases, documents, and the internet. These systems are foundational to technologies including search engines, virtual assistants, and automated customer support, demonstrating their broad utility \cite{chen2017reading}.

To enhance the accuracy of QA systems by providing more external Knowledge, the \textit{Retrieval-Augmented Generation (RAG) }\cite{lewis2020retrieval} framework has been developed. RAG addresses these challenges by combining retrieval and generation mechanisms in a cohesive process that includes:

\begin{enumerate}
    \setlength{\itemsep}{0pt} 
    \setlength{\parskip}{0pt} 
    \item \textit{Query Formulation:} Creating a query $Q_i$ based on the user's question.
    \item \textit{Document Retrieval:} Retrieving a relevant set of documents $\{c_1, c_2, \dots, c_k\}$ from a comprehensive dataset based on $Q_i$.
    \item \textit{Context Integration for Reader} Processing the retrieved documents and the query $Q_i$ through a Language Model (LM).
    \item \textit{Answer Output:} The LM integrates the inputs and generates the answer \\ $A_i = \text{LM}(R(Q_i,D),Q_i)$.
\end{enumerate}

This approach has gained significant attention in recent years due to its ability to handle open-domain questions and leverage vast amounts of unstructured text data. The retriever in RAG can be implemented using various techniques, such as BM25 \cite{robertson2009probabilistic}, dense passage retrieval with BERT \cite{devlin2019bert}, or sentence transformers \cite{reimers2019sentence}. The generator, on the other hand, is typically a large pre-trained language model like T5 \cite{raffel2020exploring}, FLAN-T5 \cite{flan_t5}, or GPT \cite{radford2019language}.

However, the effectiveness of traditional RAG systems is often hampered by biases such as positional bias, where information from initially retrieved documents is overly prioritized, potentially overlooking more relevant information in lower-ranked documents. This can lead to suboptimal and inaccurate responses, particularly evident in the following example:
\begin{quote}
\textit{Q: What was Hero of Alexandria's nationality?}\\
\textit{A: Polish}
\end{quote}
This incorrect answer reflects the system's focus on irrelevant information about famous individuals from Warsaw, Poland, rather than addressing the factual historical context of Hero of Alexandria. This error illustrates the system's propensity to latch onto irrelevant data, thereby failing to deliver accurate responses based on factual relevance \cite{zhan2021desirable}.


Alongside the above issue, current RAG systems face challenges such as computational inefficiencies and the inability to effectively prioritize relevant information from a large set of documents, which are notable issues that degrade performance. Computational inefficiencies arise from the need to process and re-rank a large number of retrieved documents, leading to slower response times and increased resource consumption. Additionally, the inability to effectively prioritize relevant information from the retrieved documents can result in the generation of answers that are not well-supported by the most relevant facts, leading to inaccurate or incomplete responses.


To address these limitations, we introduce \textit{Logit Ranked Retriever Ensemble (LoRE)}, a novel framework that leverages diverse retrieval methods to mitigate biases and enhance QA accuracy. LoRE uses BM25 and sentence transformers with FAISS \cite{johnson2019billion} indexing to improve retrieval diversity and relevance. It introduces a logit-based ranking mechanism that integrates retrieval ranks with confidence scores from an LLM like T5, refining information prioritization. This ranking mechanism refines the prioritization of information, ensuring that the most relevant and substantiated data influences the response generation. This innovative approach not only enhances the accuracy of answers but also reduces the computational overhead typically associated with re-ranking processes in existing RAG systems.

Through rigorous testing on benchmark datasets such as NarrativeQA \cite{kovcisky2018narrativeqa} and SQuAD \cite{rajpurkar2016squad}, LoRE has demonstrated substantial improvements over existing methods, achieving superior performance in both exact match and F1 scores. These results confirm that LoRE effectively addresses the key challenges in retrieval-based QA, providing a more reliable and efficient framework for handling complex QA tasks.

In summary, the proposed LoRE framework significantly advances retrieval-based QA by mitigating biases, enhancing retrieval diversity, and optimizing computational efficiency through diverse retrievers and logit-based answer ranking. This work overcomes existing system limitations and contributes robust solutions to the complex challenges in question answering within the broader field of natural language processing.


\section{Related Work}
\label{sec:related_work}

The development of retrieval-augmented LLMs has advanced significantly, focusing on improving the synergy between retrieval and generation processes. \citet{shao2023iter} introduced \textbf{ITER-RETGEN}, which iteratively enhances retrieval and generation for tasks like multi-hop QA and commonsense reasoning. It alternates between retrieval-augmented generation and generation-augmented retrieval, refining each step with the previous output. But, it faces challenges such as suboptimal utilization of retrieved information, with about 20\% of contexts lacking actual answers, leading to inaccuracies or hallucinations due to positional bias and non-optimized iterative retrieval.

Addressing similar issues, \citet{sawarkar2024blended} proposed \textbf{Blended RAG}, which incorporates semantic search and hybrid query-based retrievers to enhance the accuracy and relevancy of retrieved information. By using a combination of keyword-based, vector-based, and semantic-based searches, Blended RAG aims to mitigate the retrieval limitations observed in ITER-RETGEN by ensuring that more relevant documents are retrieved, thus improving the overall effectiveness of the RAG system. However, the complexity and computational intensity of implementing multiple retrieval strategies, as well as scalability concerns in real-world applications, remain significant challenges for Blended RAG.

Another approach by \citet{ma2023rewrite} focused on \textbf{Query Rewriting} within retrieval-augmented settings. Their "Rewrite-Retrieve-Read" framework aims to adapt the search query itself to better align with the information needs, thereby enhancing the relevancy and effectiveness of retrieval. This method reduces the dependency on extensive, domain-specific datasets by refining queries to work with existing open-domain data. Nonetheless, the system's reliance on high-quality training data and its limited ability to generalize across diverse datasets highlight persistent issues in the field.

\citet{cheng2024lift} introduced \textbf{Lift Yourself Up}, a retrieval-augmented text generation model using self-memory and external metrics like ROUGE/BLEU for more reliable and contextually relevant generation.


These studies highlight the need to refine retrieval and generation integration, addressing challenges like dynamic retrieval contexts, iterative process optimization, and scalability. Our work introduces a novel approach leveraging logit-based answer ranking and an ensemble of retrievers to mitigate positional bias and improve QA accuracy and relevance.



\section{Methodology}
\label{sec:methodology}

Our logit-based approach uses logit scores from LLMs, removing the need for expensive re-ranking steps. This makes it suitable for large-scale datasets and varied needs.
By combining an ensemble of retrievers, context iteration, and weighted logit rank scoring, we effectively address positional bias and diverse information needs.

\subsection{Ensemble of Retrievers}
\begin{figure}[ht]
\centering
\includegraphics[width=\linewidth]{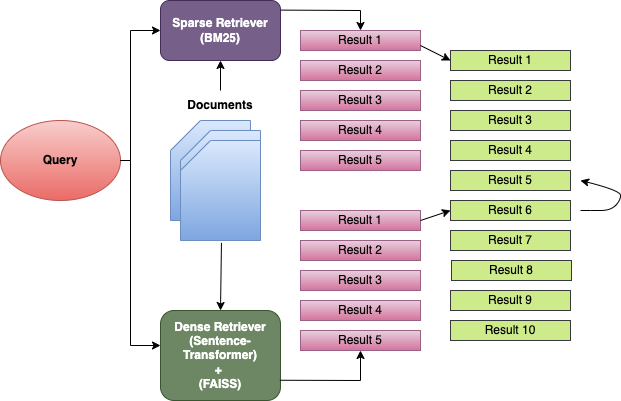}
\caption{Ensemble of Retrievers}
\label{fig:ensemble}
\end{figure}

Our approach adopts an ensemble strategy to enhance the robustness and diversity of the retrieved documents, which is crucial for handling varied and nuanced queries effectively. The ensemble consists of two main types of retrievers: a vector index retriever and a sparse n-grams index (BM25) retriever. (Fig \ref{fig:ensemble})

\vspace{-4mm}
\subsubsection{Vector Index Retrieval}
The vector index retriever uses machine learning models to map queries and documents into a high-dimensional vector space, assessing documents based on semantic similarities. This is effective for context-rich queries. We enhance retrieval efficiency with FAISS (Facebook AI Similarity Search) indexing for speedy retrieval in large datasets.

The similarity between a query \( q \) and a document \( d \) is measured using cosine similarity:
\[ \text{similarity}(q, d) = \frac{q \cdot d}{\|q\| \|d\|} \]

\begin{figure*}[ht]
\centering
\includegraphics[width=0.7\textwidth]{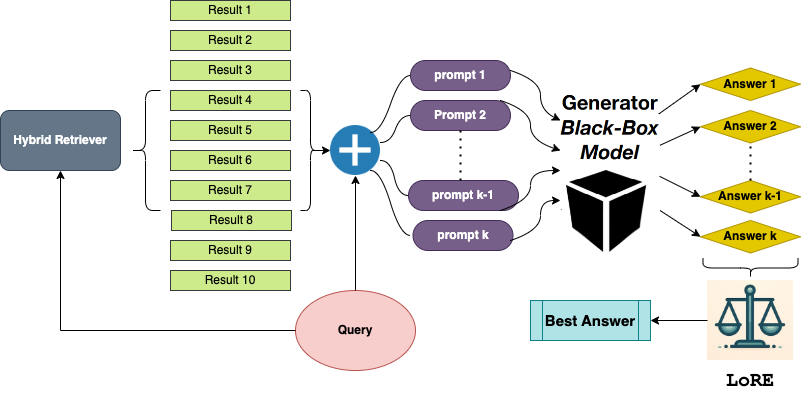}
\caption{Query and Context Interaction using T5 Model}
\label{fig:querycontext}
\end{figure*}

\subsubsection{Sparse N-grams Index Retrieval (BM25)}
The BM25 retriever operates on a different principle, utilizing the Bag of Words model to compute scores based on the frequency of query terms in documents and the inverse document frequency of the terms across a large corpus.
The BM25 score between a query \( q \) and a document \( d \) is

{\small
\[ \text{score}(q, d) = \sum_{i=1}^{n} \text{IDF}(t_i) \cdot \frac{f(t_i, d) \cdot (k_1 + 1)}{f(t_i, d) + k_1 \cdot (1 - b + b \cdot \frac{|d|}{\text{avgdl}})} \]
}

\subsubsection{Fusion of Retrieved Results}

To leverage the strengths of both retrieval methods, we employ Reciprocal Rank Fusion (RRF) \cite{cormack2009reciprocal} to combine the results. RRF is effective in integrating the diverse results produced by different retrieval models, thereby improving overall retrieval performance. The RRF score for a document \(d\) is computed as:
\[
\text{RRF}(d) = \sum_{i=1}^k \frac{1}{\text{rank}_i(d) + k}
\]
where \(k\) is a constant, typically set to 60, and \(\text{rank}_i(d)\) is the rank of document \(d\) in the \(i\)-th retrieval list.

By summing these scores across different ranked lists (vector index and BM25 retrievers), we generate a combined list sorted by RRF scores. This fusion approach enhances retrieval effectiveness by leveraging the strengths of individual retrievers.

\subsection{Query and Context Processing}
After initial retrieval, selected documents undergo granular analysis to ensure contextual relevance. We leverage the T5 large model to generate precise answers, as illustrated in Figure \ref{fig:querycontext}.

\subsubsection{Segmentation into Context Chunks}
Our methodology processes the entire content of the top \( k \) retrieved documents rather than traditional segmentation. Each context, likely containing answers, is paired with the query and analyzed by the T5 large model for comprehensive understanding.

\subsubsection{Answer Generation for Context-Query Pair with T5 Large Model}
The T5 large model, a state-of-the-art transformer-based machine learning model trained across a variety of natural language processing tasks, processes these contexts. Capable of generating human-like text, the T5 model receives a concatenated string of the query and the entire context as input, which it uses to generate potential answers. This method leverages the model's ability to discern and respond to the nuanced features within the text, producing responses that are not only relevant but deeply enriched with context. This seamless integration ensures that the integrity and narrative flow of the original document are maintained, enhancing the relevance and accuracy of the response.

Let \( C_i \) represent an individual context chunk from the top \( k \) retrieved contexts, and let \( Q \) be the query. The T5 model processes these inputs to generate an answer \( A_i \) as follows: \( \text{\(A_i \)} = \text{T5}(Q \| C_i) \), where \( \| \) denotes concatenation. This formula encapsulates how the model integrates and transforms the input query and context into a coherent answer.

By generating answers for each individual context-query pair, the methodology ensures that every potential piece of relevant information is considered. This approach mitigates the risk of missing critical information that could be pivotal in answering the query accurately.


\subsection{Logit Evaluation for Answer Validation}
To ensure the reliability and accuracy of the answers generated by the T5 model, we employ a meticulous evaluation process based on the logits produced by the model. Logits are the raw output scores from the neural network's final layer before conversion into probabilities through a softmax function.

\subsubsection{Analysis of Logits}
Logit probabilities indicate the model's confidence in its predictions. After the T5 model generates answers, we extract and analyze these logit scores to assess the model’s confidence.

High logit scores do not always ensure factual accuracy, as highlighted by \citet{cheng2024lift} and \citet{meister2020beam}. To enhance response reliability, LoRE integrates several strategies: an ensemble of diverse retrievers (BM25 and sentence transformers with FAISS), a logit-based ranking mechanism combining logit scores with retrieval ranks, and context rank integration to prioritize relevant answers. These methods minimize hallucinations and improve answer accuracy.

\subsubsection{Softmax on Logits}
The probabilities \( P \) of each token being the correct part of the answer are computed from the logits \( L \) using the softmax function:
\[ P(i) = \frac{e^{L_i}}{\sum_{j} e^{L_j}} \]
where \( L_i \) is the logit for the i-th token. The mean score \( \text{Mean Score} \) is then the average of these probabilities i.e, \textit{primary score}:
\[ \text{Mean Score} = \frac{1}{n} \sum_{i=1}^{n} P(i) \]

\subsubsection{Context Rank Integration}
The context rank i.e, \textit{secondary score} assesses the relevance of each context chunk based on its original position from the retrieval fusion phase. This ranking is essential when generated answers have similar probability scores, as it helps to refine the decision-making process. In implementation, each context is ranked by its Reciprocal Rank Fusion (RRF) score, and this rank is inversely applied in our evaluation formula. Higher-ranked contexts, indicating greater relevance, significantly influence the weighted score. This method enhances the accuracy of selecting the most pertinent answer by factoring in the relevance assessed during the initial retrieval.

\subsubsection{LoR score}
\label{LoR score}
The LoR score integrates both the mean score and the context rank, balancing the model's confidence with the intrinsic relevance of the context. This dual consideration ensures that answers are not only probable according to the model but are also derived from the most relevant sections of the document. 
 The weighted score, \( LoR \), is calculated using the formula below, where $w_1$ and $w_2$ are empirically set to 0.8 and 0.2:
\vspace{-5pt}
\[ LoR = (w_1 \times \text{Mean Score}) + (w_2 \times \frac{1}{\text{Context Rank}}) \]
\vspace{-15pt}
\begin{figure}[ht]
\centering
\includegraphics[width=0.7\linewidth]{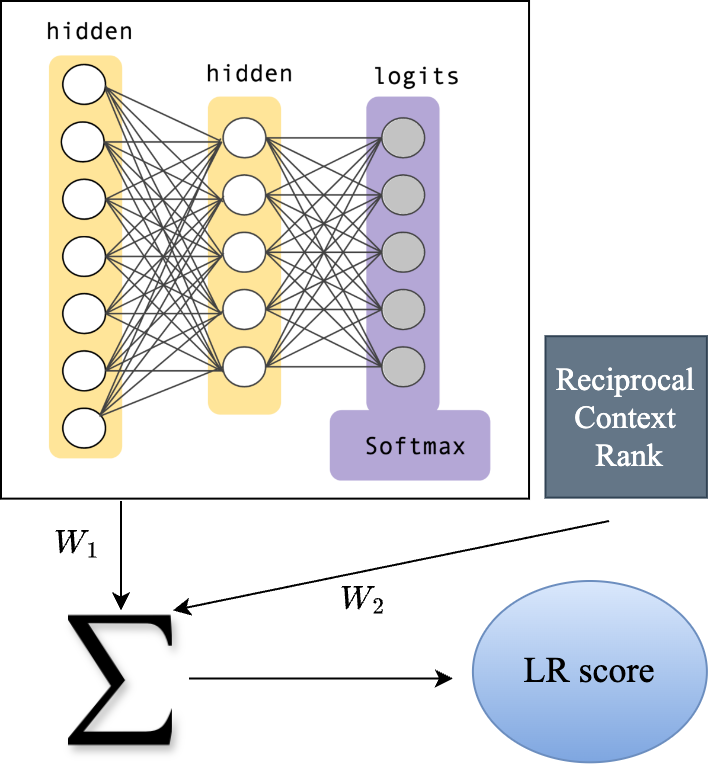}
\caption{Logit Evaluation and Context Rank Integration}
\label{fig:logitcontext}
\end{figure}

This logit evaluation framework is essential for maintaining high standards of answer validation, ensuring that the answers not only sound correct but are substantiated by both model confidence and contextual relevance. An illustration of this process is depicted in Figure \ref{fig:logitcontext}.



\section{Experimentation}
\label{sec:experiments}
This section evaluates our models from two perspectives: (1) The ability to mitigate positional bias in open-domain question answering systems; (2) The effectiveness of our ensemble retriever methods in improving the accuracy and relevance of the answers provided.

\subsection{Experimental Setup}

\subsubsection{Datasets}
For our experiments, we utilized two datasets to test the generality of our proposed approaches in addressing positional bias:

\textit{Stanford Question Answering Dataset (SQuAD):} Comprises over 100,000 question-answer pairs based on passages from Wikipedia articles..

\textit{Narrative Question Answering (NarrativeQA):} Has over 3,460 question-answer pairs, the NarrativeQA dataset provides a unique challenge by basing questions on the summaries and full texts of 767 literary works and movie scripts.We obtain the summaries of the retrieved contexts and rank them based on our methodologies to achieve our answer.

\subsubsection{Baselines}
We benchmark our model against three baseline setups to illustrate the improvements our approach provides:
\begin{itemize}
    \item \textit{BM25 and T5 Large:} Utilizes the BM25 algorithm for document retrieval combined with the T5 large model for generating answers.
    \item \textit{VectorDB (Sentence Transformer + FAISS) and T5 Large:} Employs Sentence Transformer for embedding generation, FAISS for efficient similarity search, and T5 Large for answer generation.
    \item \textit{Ensemble of BM25 and Sentence Transformer with T5 Large: }This setup integrates both BM25 and Sentence Transformer approaches to maximize the relevance and accuracy of the retrieved documents.

\end{itemize}



The model proposed in the Blended RAG paper utilizes 11B parameters, whereas LoRE is able to accomplish an improvement of 4\% with the use of just 770M. This is a decrease of approximately 10.3 billion parameters.

\subsubsection{Metrics}
We assessed our models using the following evaluation metrics: \textbf{ROUGE-L:} Measures the longest common subsequence of words. \textbf{Exact Match (EM) Score:} Checks if the predicted answer matches the ground truth exactly. \textbf{F1 Score:} Harmonic mean of precision and recall, important for understanding the accuracy and relevance of the retrieved answers.

\begin{table*}
  \centering
  \small
  \begin{tabular}{lccc|ccc}
    \hline
    \textbf{Model} & \multicolumn{3}{c|}{\textbf{SQUAD Dataset}} & \multicolumn{3}{c}{\textbf{NarrativeQA Dataset}} \\
    & \textbf{EM} & \textbf{F1} & \textbf{Rouge-L} & \textbf{EM} & \textbf{F1} & \textbf{Rouge-L} \\
    \hline
    RAG original & 28.12\% & 39.42\% & - & - & - & - \\
    RAG end2end & 40.02\% & 52.63\% & - & - & - & - \\
    Blended RAG & 57.63\% & 68.40\% & - & - & - & - \\
    Naive - BM25 & 19.41\% & 34.78\% & 31.90\% & 0.00\% & 3.33\% & 2.34\% \\
    Naive - FAISS & 26.40\% & 42.80\% & 39.40\% & 0.00\% & 3.23\% & 2.32\% \\
    Naive - Ensemble & 38.62\% & 54.32\% & 50.30\% & 0.00\% & 3.29\% & 2.33\% \\
    LoRE - BM25 & 56.66\% & 63.91\% & 59.30\% & \textbf{31.99\%} & \textbf{48.92\%} & \textbf{41.39\%} \\
    LoRE - FAISS & 57.28\% & 65.19\% & 60.10\% & 27.68\% & 42.55\% & 35.41\% \\
    LoRE - Ensemble & \textbf{61.45\%} & \textbf{69.27\%} & \textbf{64.80\%} & 31.23\% & 47.92\% & 40.37\% \\
    \hline
  \end{tabular}
  \caption{Performance metrics on SQUAD and NarrativeQA Datasets}
\end{table*}
\subsection{Experiment Results}
\label{sec:results}
\subsubsection{Quantitative Results}
Our results indicated significant improvements over the baselines, as demonstrated in our experiments with the SQuAD dataset:

\begin{itemize}
    \item Ensemble + T5 Large with our logits approach achieved a ROUGE-L score of 64.8\%, EM Score of 61.45\%, and F1 Score of 69.27\%, which represents improvements of 14.5\%, 22.83\%, and 14.95\% over the respective baseline scores.
\end{itemize}

\noindent For the NarrativeQA dataset, preliminary results shows that Ensemble + T5 Large with our logits approach achieved a ROUGE-L score of 48.8\%, EM Score of 31.23\%, and F1 Score of 47.92\%, which represents improvements of 14.5\%, 31.23\%, and 44.60\% over the respective baseline scores.

Kindly note that the BM25 algorithm achieved superior performance compared to the ensemble approach in this particular instance. This can be attributed to the considerably greater length of the summaries within the NQA dataset. Consequently, the ensemble method encountered challenges in effectively comprehending and capturing the contextual nuances present in the more extensive text, thereby resulting in a lower score relative to the BM25 algorithm.
\subsubsection{Qualitative Results}

We conducted a qualitative assessment of LoRE's performance by comparing its generated answers to those from the traditional blended RAG model. LoRE demonstrated superior accuracy in identifying and prioritizing relevant answers, as evidenced by its enhanced performance on both SQUAD and NarrativeQA datasets. LoRE significantly improved answer relevancy and reduced hallucinations compared to the blended RAG model.

For instance, in response to the query, "What petroleum company was a Super Bowl sponsor?", LoRE accurately identified "Chevron", whereas the base ensemble model incorrectly predicted "neo-maeli". This illustrates LoRE's effective use of context rank and probability to generate precise answers. Results, summarized in Tables 2 and 3, show that LoRE consistently outperformed the baseline, particularly in scenarios where the baseline model produced less relevant or hallucinated answers.

\section{Conclusion}
\label{sec:conclusion}
\noindent This paper proposed LoRE, a novel approach that combines an ensemble of diverse retrievers with a logit-based ranking algorithm to mitigate positional bias in retrieval-based question-answering systems. By integrating logit scores from a large language model with retrieval ranks, LoRE effectively assesses answer reliability and selects the most appropriate one.
\noindent Experiments on multiple benchmarks demonstrate LoRE's significant performance gains over existing retrieval-based methods in terms of exact match and F1 scores. Qualitative analysis further confirms LoRE's ability to generate more relevant and accurate answers, especially for complex queries requiring deep contextual understanding.


\section{Limitations and Future Work}
\label{sec:limitations}
\noindent

The accuracy of logit scores for evaluating answer relevance is highly dependent on the diversity and representativeness of the training data, furthermore, the model is dependent on the performance of the retriever for the initialization.

Future work should aim to generalize the application of logit evaluations to other generative models beyond T5 Large, conducting systematic studies to understand their logit processing and generation. Additionally, we are currently calling the model \textit{k} times to receive \textit{k} results, although we attempt to offset this by using a much smaller model, our method can be optimized by engineering the prompt to generate the outputs all at once.

In the LoR score calculation, $w_1$ and $w_2$ were chosen empirically, but making these weights trainable parameters could lead to a more advanced and adaptive system for ranking answers. By enabling the model to learn and adjust these values from data, the system could better adapt to varying contexts, thereby enhancing the overall effectiveness of the LoRE approach.

\bibliographystyle{ACM-Reference-Format}
\bibliography{camready-sigconf}

\clearpage

\section{Appendix}
\label{sec:appendix}

\begin{table}[ht]
\centering
\label{squad_qual}
\centering

\small 
\setlength{\tabcolsep}{5pt} 
\renewcommand{\arraystretch}{1.2} 
\begin{tabular}{|p{2.5cm}|p{1.5cm}|p{2cm}|p{2.5cm}|p{2cm}|p{1.5cm}|p{1.5cm}|} 
\hline
\textbf{Baseline query} & \textbf{True answer} & \textbf{Baseline predicted answer} & \textbf{Logit predicted answer} & \textbf{Logit probabilities} & \textbf{Logit predicted answer} & \textbf{Context answer rank} \\ 
\hline
 Question: Who is the quarterback for the Panthers?; Context: the panthers offense, which led the nfl in scoring (500 points), was loaded with talent, boasting six pro bowl selections ... & Cam Newton & denver broncos won the super bowl 50, becoming the first head coach to win' & "['cam newton', 'newton', 'peyton manning', 'manning', 'newton', 'cam newton', 'Jeremy Lamb', 'Ryan', 'manning']" & [0.995, 0.973, 0.997, 0.928, 0.986, 0.991, 0.311, 0.189, 0.722] & cam newton' & 0 \\
\hline
 Question: When was the last Super Bowl in California?; Context: super bowl 50 was an american football game to determine the champion of the national football league (nfl) for the 2015 season. the american football ... & 2003 & denver broncos defeated the national football conference champion carolina panthers 24' & "['february 7, 2016', 'super bowl xxxiii', '2003', 'prior to super bowl 50', '1985', '1999', '2001', 'february 1, 2016', '1999', '1995']" & [0.794, 0.931, 0.984, 0.959, 0.974, 0.551, 0.694, 0.948, 0.531, 0.671] & 2003 & 2  \\
\hline
 Question: How many points did the Broncos defense allow their opponents to get?;Context: the panthers seemed primed to score on their opening drive of the second half when newton completed a 45-yard pass to ted ... & 296 & denver broncos defense ranked first in the nfl yards allowed (4' & "['a total of 62', 'no', 'a total of 62', '10', 'a total of 63', 'a sizeable population of huguenot descent lived in the british colonies', 'a technical defense may enhance the chances for acquittal but make for more boring', '296']" & [0.366, 0.531, 0.390, 0.671, 0.412, 0.869, 0.890, 0.944] & 296 & 7 \\
\hline
\end{tabular}
\caption{LoRE Model performance on SQUAD Dataset}
\end{table}

\clearpage

\begin{table}[ht]
\label{narrativeqa_qual}

\centering

\small 
\setlength{\tabcolsep}{5pt} 
\renewcommand{\arraystretch}{1.2} 
\begin{tabular}{|p{2.5cm}|p{1cm}|p{1.5cm}|p{3cm}|p{2cm}|p{1.5cm}|p{1.5cm}|} 
\hline
\textbf{Baseline query} & \textbf{True answer} & \textbf{Baseline predicted answer} & \textbf{Logit predicted answer} & \textbf{Logit probabilities} & \textbf{Logit predicted answer} & \textbf{Context answer rank} \\ 
\hline
 Question: who challenges the courtiers to court compliment competition?; Context: the play begins with three pages disputing over the black cloak usually worn by the ... & asotus & elizabeth cole (elwes) is the daughter of ' & ['asotus', 'RUTH', 'fletcher reede (carrey) loves his son max (cooper', 'VALJENNIE', "mrs. cheveley, an enemy of lady chiltern'", 'drago bludvist', 'DICY', 'wolfe challenges murney to a wrestling match in front of the entire school and easily', 'JOE CLARKE', 'death', 'domino harvey, a bounty hunter, has been arrested by the', 'court reynolds', 'elizabeth i of england (cate blanchett)', 'TURKISH RIVERS', 'truman capote', 'ada clare is pregnant and richard is in the last stages of tuber', 'CHARLES VEAL jr.']' & [0.982, 0.652, 0.940, 0.683, 0.917, 0.901, 0.764, 0.902, 0.798, 0.911, 0.952, 0.895, 0.948, 0.559, 0.835, 0.632, 0.839] & asotus' & 1 \\
\hline
 Question: What color is the slime the Ghostbusters find underground?;Context: after saving new york city from the demi-god gozer, the ghostbustersâ peter venkman, egon spengler, ray stantz, and winston ... & Pink. & and a his remembrance of his father and brother. ' & [\'green\', \'jimmy gator is dying of cancer; he has only a few months\', \'pink\', \'blue\', \'<unk>\', \'black\', \'blue\', \'elizabeth i of england (cate blanchett)\', \'rainbow\', \'green\', \'blue\', \'blue\', "wolfe\'s commanding officer, captain bill fawcett, is", \'blue\', \'green\', \'indiana jones and his partner george "mac" mchale\', \'a dark brown slime\', \'green\']' & [0.581, 0.860, 0.988, 0.585, 0.597, 0.904, 0.571, 0.909, 0.597, 0.598, 0.615, 0.606, 0.844, 0.604, 0.584, 0.905, 0.466, 0.606] & pink & 2 \\
\hline
 Question: Which page performs the dialogue?; Context: dorian gray is the subject of a full-length portrait in oil by basil hallward, an artist who is impressed and infatuated by dorian\'s beauty; he believes that dorian’s beauty is ... & Anaides & he is a deemster, and he is a sa' & ["dorian gray\'s picture of dorian gray", \'jim kurring\', \'spitfire\', \'anaides\', "wolfe\'s commanding officer, captain bill fawcett, is", \'page\', \'thomas jerome newton\', \'canto 3\', \'the page\', \'page\', \'mulder and scully have been assigned to other projects since the closure of the\', \'liza kemp is an 18-year-old factory worker and the youngest of\', \'armageddon 2419 a.d.\', \'mrs. cheveley is an enemy of lady chiltern and\', \'elaine\', \'gilbert markham narrates how a mysterious widow, mr\', \'page 62\', \'page\', \'page\', \'page\']' & [0.723, 0.769, 0.854, 0.961, 0.890, 0.383, 0.868, 0.741, 0.336, 0.210, 0.927, 0.918, 0.938, 0.793, 0.830, 0.924, 0.342, 0.522, 0.335, 0.360] & anaides & 3  \\
\hline

\end{tabular}
\caption{LoRE Model performance on NarrativeQA}
\end{table}

\clearpage

\begin{figure*}[ht]
\subsection{Motivation for the use of Logits}
\label{logits_motivation}
    \centering
    \begin{minipage}[t]{0.45\linewidth}
        \centering
        \includegraphics[width=\linewidth]{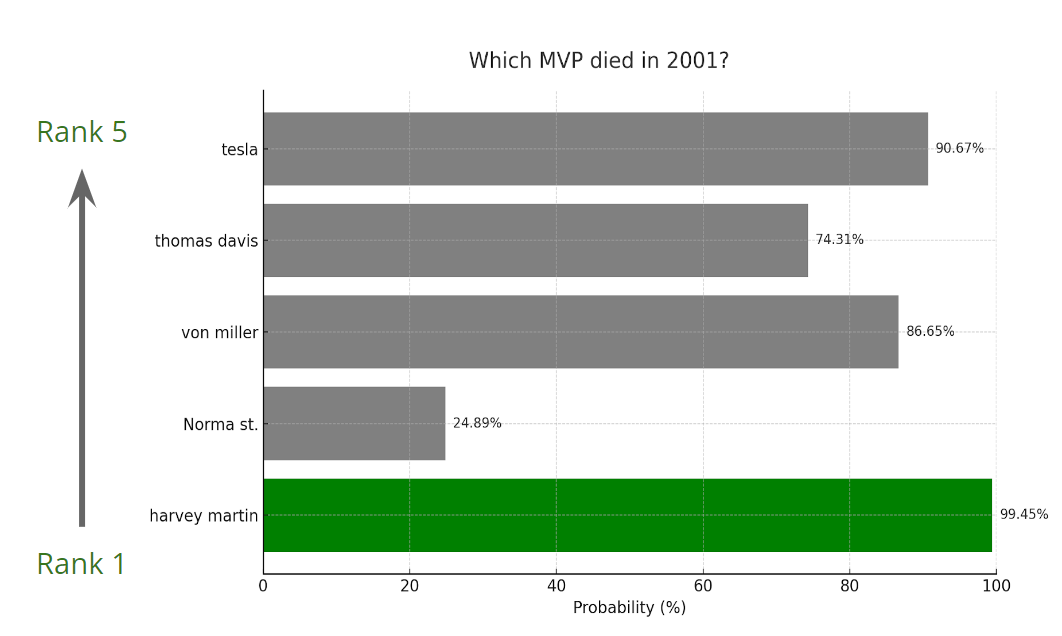}
        \caption{The one with the highest probability is the correct answer and was also ranked first}
        \label{fig:slide35}
    \end{minipage}
    \hfill
    \begin{minipage}[t]{0.45\linewidth}
        \centering
        \includegraphics[width=\linewidth]{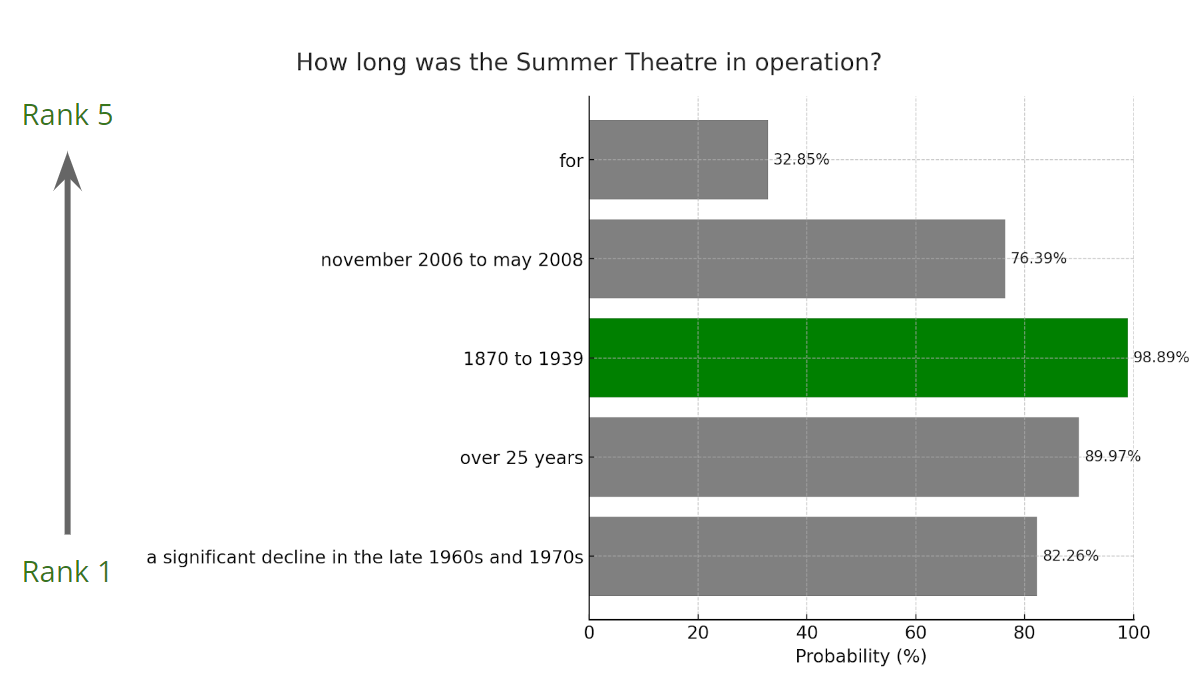}
        \caption{The one with the highest probability was the correct answer even though it was not ranked first}
        \label{fig:slide36}
    \end{minipage}
    
    \medskip
    \vspace{4cm}
    
    \begin{minipage}[t]{0.45\linewidth}
        \centering
        \includegraphics[width=\linewidth]{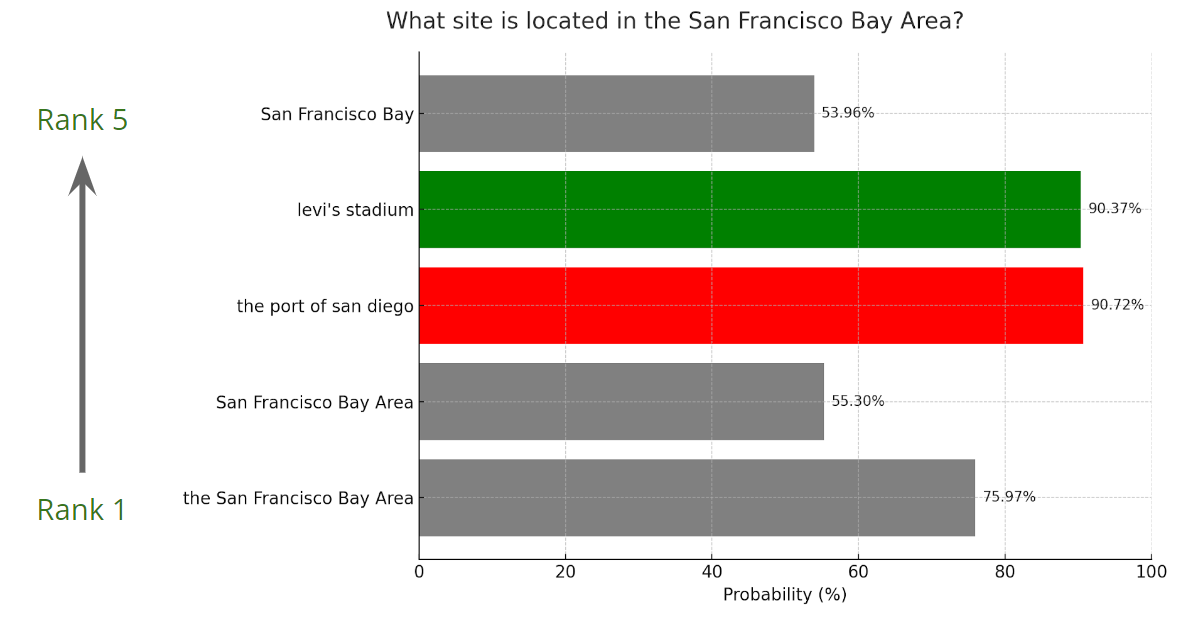}
        \caption{While logits may not always provide the correct answer, it will reduce hallucinations, as a lower probability score indicates irrelevant responses}
        \label{fig:slide37}
    \end{minipage}
    \hfill
    \begin{minipage}[t]{0.45\linewidth}
        \centering
        \includegraphics[width=\linewidth]{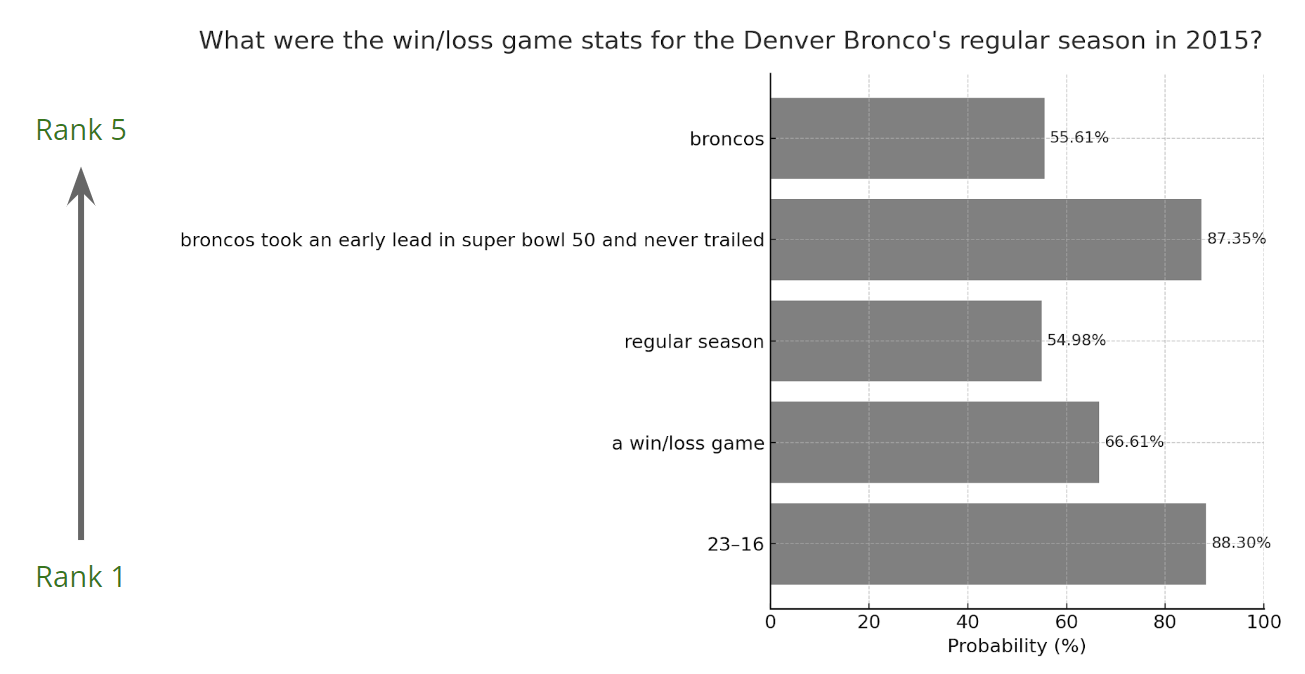}
        \caption{If the answer is not present in the context, this method would also not work, but it would give the highest probability to that which has the most similar format to the true correct answer}
        \label{fig:slide38}
    \end{minipage}
    \label{fig:methods_comparison}
\end{figure*}

\begin{figure*}[ht]
\begin{minipage}[t]{\linewidth}
\centering
\includegraphics[width=\linewidth]{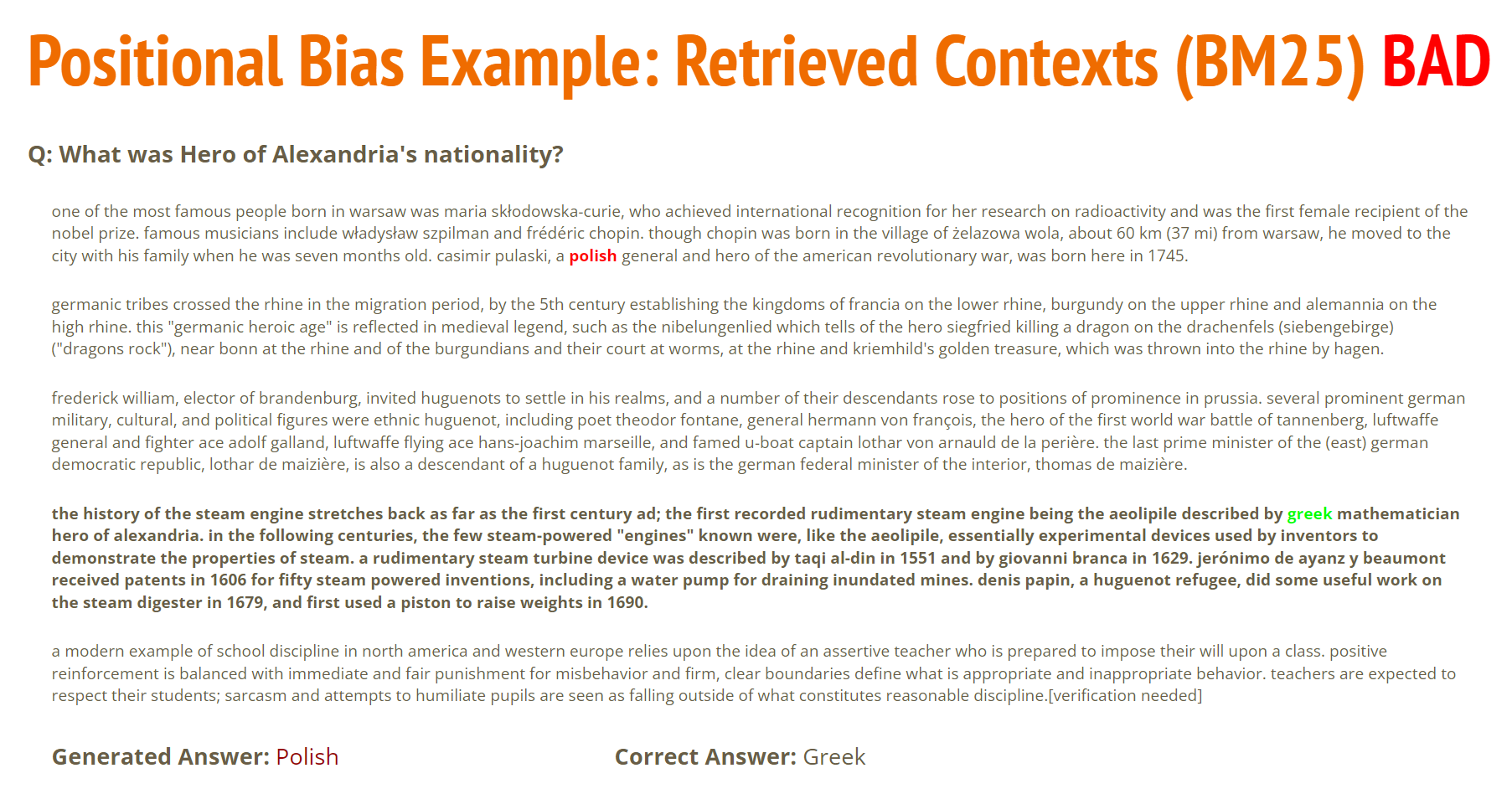}
\caption{While logits may not always provide the correct answer, it will reduce hallucinations, as a lower probability score indicates irrelevant responses}
\label{fig:bad}
\end{minipage}
\begin{minipage}[t]{\linewidth}
\centering
\vspace{2cm}
\includegraphics[width=\linewidth]{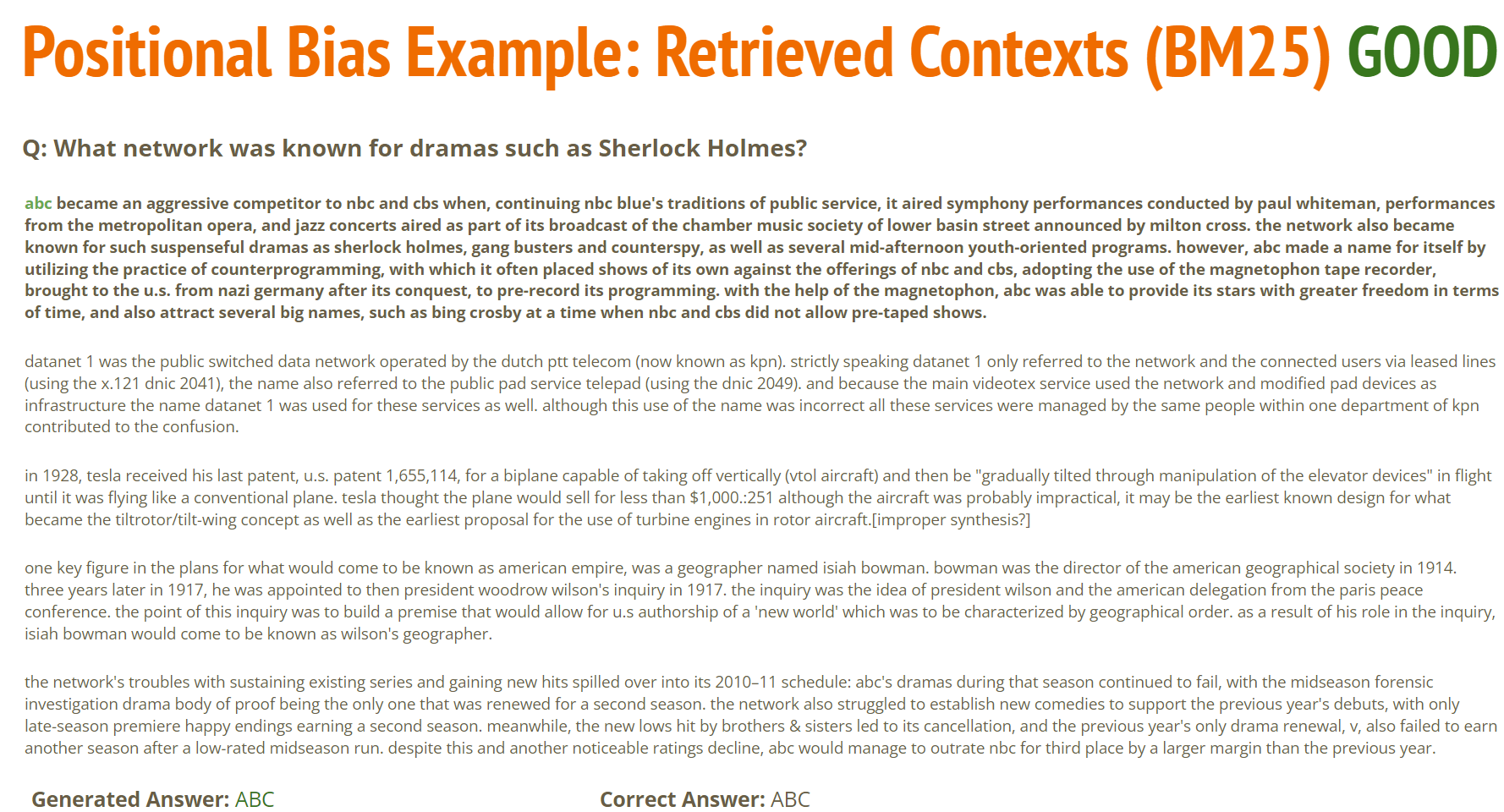}
\caption{If the answer is not present in the context, this method would also not work, but it would give the highest probability to that which has the most similar format to the true correct answer}
\label{fig:good}
\end{minipage}
\end{figure*}
\clearpage

\newpage

\subsection{Reciprocal Rank Fusion Example}

Suppose you have the following ranked lists from two retrieval methods:

\begin{itemize}
    \item \textbf{BM25}: [D1, D2, D3, D4]
    \item \textbf{Sentence Transformer}: [D3, D1, D5, D2]
\end{itemize}

Let's assume \(k = 60\). Here’s how we combine them using Reciprocal Rank Fusion (RRF):

\subsection*{Assign Ranks}

\begin{table}[h]
\centering
\begin{tabular}{ccc}
\toprule
\textbf{Document} & \textbf{Rank (BM25)} & \textbf{Rank (Sentence Transformer)} \\
\midrule
D1 & 1 & 2 \\
D2 & 2 & 4 \\
D3 & 3 & 1 \\
D4 & 4 & $\infty$ \\
D5 & $\infty$ & 3 \\
\bottomrule
\end{tabular}
\caption{Assigned Ranks}
\end{table}

\subsection*{Calculate RRF Scores}

\[
\text{RRF}(d) = \sum_{i=1}^k \frac{1}{\text{rank}_i(d) + k}
\]

\[
\text{RRF}(D1) = \frac{1}{1 + 60} + \frac{1}{2 + 60} = \frac{1}{61} + \frac{1}{62} \approx 0.03226
\]

\[
\text{RRF}(D2) = \frac{1}{2 + 60} + \frac{1}{4 + 60} = \frac{1}{62} + \frac{1}{64} \approx 0.01613 + 0.01587 = 0.03200
\]

\[
\text{RRF}(D3) = \frac{1}{3 + 60} + \frac{1}{1 + 60} = \frac{1}{63} + \frac{1}{61} \approx 0.01587 + 0.01613 = 0.03200
\]

\[
\text{RRF}(D4) = \frac{1}{4 + 60} + \frac{1}{\infty} = \frac{1}{64} + 0 \approx 0.01587
\]

\[
\text{RRF}(D5) = \frac{1}{\infty} + \frac{1}{3 + 60} = 0 + \frac{1}{63} \approx 0.01587
\]

\subsection*{Sum the Scores and Sort}

\begin{table}[h]
\centering
\begin{tabular}{cc}
\toprule
\textbf{Document} & \textbf{RRF Score} \\
\midrule
D1 & 0.03226 \\
D2 & 0.03200 \\
D3 & 0.03200 \\
D4 & 0.01587 \\
D5 & 0.01587 \\
\bottomrule
\end{tabular}
\caption{RRF Scores}
\end{table}

Thus, the combined list, sorted by RRF scores, is: [D1, D2, D3, D4, D5].

\end{document}